\documentclass{article}

\usepackage{microtype}
\usepackage{graphicx}
\usepackage{booktabs} 

\usepackage{hyperref}

\usepackage{microtype}
\usepackage{graphicx}
\usepackage{booktabs} 
\usepackage{hyperref}

\usepackage[accepted]{icml2019}

\usepackage{times}
\usepackage{epsfig}
\usepackage{graphicx}
\usepackage{amsmath}
\usepackage{amssymb}
\usepackage[utf8]{inputenc}
\usepackage{subcaption}
\usepackage{tumabbrev}

\usepackage{appendix}

\usepackage[capitalize]{cleveref}

\usepackage{tikz}
\usepackage{pgfplots}
\usetikzlibrary{positioning, calc,arrows,arrows.meta, fit}
\usepgfplotslibrary{groupplots}
\usepgfplotslibrary{fillbetween}
\usepgfplotslibrary{statistics} 
\usepackage{xfrac}

\usetikzlibrary{shapes,snakes}

\usepackage{tumcolors}
\usepackage{tummath}
\newcommand{\yhat}{\hat{\V{y}}}
\newcommand{\ycorrect}{\hat{y}^+}
\newcommand{\thetadelta}{\V{\Theta}_\delta}
\newcommand{\biasdelta}{b_\delta}
\newcommand{\biasclass}{\V{b}_\text{c}}
\newcommand{\thetaclass}{\V{\Theta}_\text{c}}

\newcommand{\fclass}{f_\text{c}}
\newcommand{\fdelta}{f_\delta}
\newcommand{\ffeat}{f_\text{feat}}

\newcommand{\xuptot}{\M{X}_{\rightarrow t}} 
\newcommand{\deltauptot}{\delta_{\rightarrow t}} 
\newcommand{\tstop}{\ensuremath{t_\text{stop}}}
\newcommand{\meantstop}{\ensuremath{\bar{t}_\text{stop}}}
\usepackage[super]{nth}
\usepackage{mathtools}

\definecolor{evalcolor}{HTML}{3F3F3F}
\definecolor{traincolor}{HTML}{B98951}
\definecolor{validcolor}{HTML}{3F4BBE}

\colorlet{colortrain}{tumblue}
\colorlet{colorinfer}{tumblack}

\colorlet{earlinesscolor}{tumblue}
\colorlet{accuracycolor}{tumorange}

\colorlet{stdcolor}{tumbluelight}
\colorlet{mediancolor}{tumorange}
\colorlet{meancolor}{tumblue}

\colorlet{b1color}{tumblack}
\colorlet{b9color}{tumblack}
\colorlet{b10color}{tumblack}

\colorlet{b2color}{tumblue}
\colorlet{b3color}{tumblue}
\colorlet{b4color}{tumblue}

\colorlet{b5color}{tumdiagramred}
\colorlet{b6color}{tumdiagramred}
\colorlet{b7color}{tumdiagramred}
\colorlet{b8color}{tumdiagramred}
\colorlet{b8Acolor}{tumdiagramred}

\colorlet{b11color}{tumorange}
\colorlet{b12color}{tumorange}

\colorlet{epsilon0color}{tumorange}
\colorlet{epsilon1color}{tumblue}
\colorlet{epsilon10color}{tumblack}

\colorlet{meadowcolor}{tumbluemedium}
\colorlet{wbarleycolor}{tumbluedark}
\colorlet{corncolor}{tumorange}
\colorlet{wheatcolor}{tumgreen}
\colorlet{sbarleycolor}{tumdiagramred}
\colorlet{clovercolor}{tumdiagramturquoise}
\colorlet{triticalecolor}{tumdiagramsand}

\tikzstyle{rnn}=[draw,circle, inner sep=.1em]
\tikzstyle{norm}=[rounded corners,draw]
\tikzstyle{annot}=[rounded corners, fill=tumblue!20]
\tikzstyle{infer}=[-stealth, shorten >=.0em, shorten <=.0em, colorinfer]
\tikzstyle{loss}=[fill=tumblue!10, rounded corners, font=\small]
\tikzstyle{grad}=[colortrain]

\newcommand{\ptoffset}{\varepsilon}

\tikzstyle{test} = [thick]
\tikzstyle{train} = [thin, dotted]

\usepackage[inline]{enumitem}
\setenumerate{label=(\roman*),itemsep=3pt,topsep=3pt}

\newcommand{\deltat}{\ensuremath{p_t}}

\usetikzlibrary{external}
\tikzexternaldisable

\usepackage[eulergreek]{sansmath}
\pgfplotsset{
	y tick label style={/pgf/number format/.cd,%
		scaled y ticks = false,
		set thousands separator={},
		fixed},
	x tick label style={/pgf/number format/.cd,%
		scaled x ticks = false,
		set decimal separator={,},
		fixed},
	tick label style = {font=\scriptsize\sansmath\sffamily},
	every axis label = {
		font=\scriptsize\sansmath\sffamily},
	every axis/.append style={
		axis lines=left, 
		enlargelimits, 
		thick},
	legend style = {font=\scriptsize\sansmath\sffamily, draw=none, rounded corners, fill opacity=.5, text opacity=1},
	label style = {font=\scriptsize\sansmath\sffamily},
	grid style={line width=.1pt, draw=gray!10},
	major grid style={line width=.2pt,draw=tumgraylight},
}

\let\tempone\itemize
\let\temptwo\enditemize
\renewenvironment{itemize}{\tempone\addtolength{\itemsep}{-.5\baselineskip}}{\temptwo}

\tikzstyle{circ} = [circle, draw=white, fill=tumblue, inner sep=1pt]
\newcommand{\fcn}{
	\begin{tikzpicture}[scale=0.2, rotate=0, baseline=-.25em, inner sep=1pt]
	\node[circ](a0) at (0,-1){};
	\node[circ](a1) at (0,0){};
	\node[circ](a2) at (0,1){};
	
	\node[circ](b0) at (1,-0.5){};
	\node[circ](b1) at (1,0.5){};
	
	\draw[-] (a0) -- (b0);
	\draw[-] (a1) -- (b0);
	\draw[-] (a2) -- (b0);
	
	\draw[-] (a0) -- (b1);
	\draw[-] (a1) -- (b1);
	\draw[-] (a2) -- (b1);
	
	\end{tikzpicture}
}

\newcommand{\hidden}[1]{
	\begin{tikzpicture}[scale=.1, baseline=-.25em]	
	\foreach \i in {1,...,#1}{
		\node[circle, draw=white, fill=tumbluelight, inner sep=1pt] at (\i,0){};
	}
	\end{tikzpicture}
}

\newcommand{\drawvector}[1]{
	\begin{tikzpicture}[scale=.1, baseline=-.25em]	
	\foreach \i in {1,...,#1}{
		\node[circ] at (\i,0){};
	}
	\end{tikzpicture}
}

\tikzstyle{druschdatum} = [thin, star,star points=3, star point ratio=0.5, inner sep=.15em, draw=tumwhite, fill=tumblue]

\newcommand{\druschdatum}{
\begin{tikzpicture}[scale=2, baseline=-.25em, inner sep=0]
\node[druschdatum, inner sep=.25em]{};
\end{tikzpicture}
}

\icmltitlerunning{Early Classification for Agricultural Monitoring from Satellite Time Series}

\begin{document}
	
	\twocolumn[
	\icmltitle{Early Classification for Agricultural Monitoring from Satellite Time Series}
	
	
	
	\icmlsetsymbol{equal}{*}
	
	\begin{icmlauthorlist}
		\icmlauthor{Marc Ru\ss{}wurm}{tum}
		\icmlauthor{Romain Tavenard}{legt,irisa}
		\icmlauthor{Sébastien Lefèvre}{ubs,irisa}
		\icmlauthor{Marco Körner}{tum}
	\end{icmlauthorlist}
	
	\icmlaffiliation{tum}{Technical University of Munich}
	\icmlaffiliation{ubs}{Université de Bretagne Sud}
	\icmlaffiliation{irisa}{Institut de Recherche en Informatique et Systèmes Aléatoires (IRISA)}
	\icmlaffiliation{legt}{LETG-Rennes}
	
	\icmlcorrespondingauthor{Marc Ru\ss{}wurm}{marc.russwurm@tum.de}

	\icmlkeywords{Machine Learning, ICML, Early Classification, Time Series, Agriculture, Sentinel 2, Crop Type Mapping, Crop Identification}
	
	\vskip 0.3in
	]
	
	
	
	\printAffiliationsAndNotice{} 
	
	\begin{abstract}
		In this work, we introduce a recently developed early classification mechanism to satellite-based agricultural monitoring.
		It augments existing classification models by an additional stopping probability based on the previously seen information.
		This mechanism is end-to-end trainable and derives its stopping decision solely from the observed satellite data.
		We show results on field parcels in central Europe where sufficient ground truth data is available for an empiric evaluation of the results with local phenological information obtained from authorities.
		We observe that the recurrent neural network outfitted with this early classification mechanism was able to distinguish the many of the crop types before the end of the vegetative period.
		Further, we associated these stopping times with evaluated ground truth information and saw that the times of classification were related to characteristic events of the observed plants' phenology.
	\end{abstract}
	
	\section{Introduction}
	
	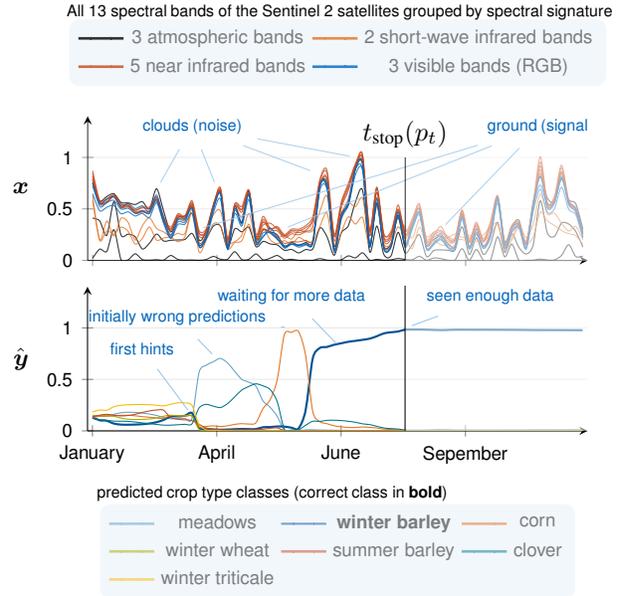
\begin{figure}[t]
		\tikzsetnextfilename{example}
\begin{tikzpicture}
	
	\tikzstyle{annot} = [font=\tiny\sffamily, text=tumblue]
	\tikzstyle{point} = [thin, tumbluelight, shorten >= .25em, shorten <= .25em]
	
	\def\tstopv{0.6285714285714286}
	\def\class{winter barley}
	
	\begin{groupplot}[
	group style={
		group name=my plots,
		group size=1 by 2,
		columns=1,
		xlabels at=edge bottom,
		xticklabels at=edge bottom,
		vertical sep=1em,
	},
	ylabel near ticks,
	ylabel style={font=\sffamily\small, rotate=-90},
	width=.48*\textwidth,
	height=3.5cm,
	axis x line=bottom,
	axis y line=left,
	enlarge x limits=0.01,
	xtick={0,0.25,0.5,0.75,1},
	xticklabels={January,April,June,Sepember,December},
	ymajorgrids,
	ymin=0, ymax=1.4
	]
	
	\nextgroupplot[thin,
		no marks,  
		ylabel={$\V{x}$},
		draw opacity=.8,
		smooth,
		legend columns=2,
		legend style={at={(.5,1.2)},anchor=south, line width=1pt, fill=tumblue!10}
		]
		 
	\addplot[b1color] table [x=t, y=B1, col sep=comma, forget plot] {images/example/input.csv};
	\addplot[b9color] table [x=t, y=B9, col sep=comma, forget plot] {images/example/input.csv};
	\addplot[b10color] table [x=t, y=B10, col sep=comma] {images/example/input.csv};
	
	\addplot[b11color] table [x=t, y=B11, col sep=comma, forget plot] {images/example/input.csv};
	\addplot[b12color] table [x=t, y=B12, col sep=comma] {images/example/input.csv};
	
	\addplot[b5color] table [x=t, y=B5, col sep=comma, forget plot] {images/example/input.csv};
	\addplot[b6color] table [x=t, y=B6, col sep=comma, forget plot] {images/example/input.csv};
	\addplot[b7color] table [x=t, y=B7, col sep=comma, forget plot] {images/example/input.csv};
	\addplot[b8color] table [x=t, y=B8, col sep=comma, forget plot] {images/example/input.csv};
	\addplot[b8Acolor] table [x=t, y=B8A, col sep=comma] {images/example/input.csv};
		
	\addplot[b2color] table [x=t, y=B2, col sep=comma, forget plot] {images/example/input.csv};
	\addplot[b3color] table [x=t, y=B3, col sep=comma, forget plot] {images/example/input.csv};
	\addplot[b4color] table [x=t, y=B4, col sep=comma] {images/example/input.csv};

	\draw[fill=white, draw=none, opacity=.5] (axis cs:\tstopv,0) rectangle (axis cs:1,1.1);
	
	\node[annot](cllab) at (axis cs:.2,1.3) {clouds (noise)};
	\draw[point] (cllab) -- (axis cs:.13,.7);
	\draw[point] (cllab) -- (axis cs:.25,.7);
	\draw[point] (cllab) -- (axis cs:.53,1);
	\draw[point] (cllab) -- (axis cs:.45,.85);
	
	\node[annot](glab) at (axis cs:.9,1.3) {ground (signal)};
	\draw[point] (glab) -- (axis cs:.38,.3);
	\draw[point] (glab) -- (axis cs:.21,.3);
	\draw[point] (glab) -- (axis cs:.7,.3);
	
	\draw (axis cs:\tstopv,0) -- (axis cs:\tstopv,1) node[above]{$\tstop(\deltat)$};

	\legend{3 atmospheric bands, 2 short-wave infrared bands, 5 near infrared bands, 3 visible bands (RGB)}

	\nextgroupplot[thin,
		smooth,
		no marks, 
		ylabel={$\yhat$},
		legend style={at={(.5,-.5)},anchor=north, line width=1pt, fill=tumblue!10},
		legend columns=3]	
	
	\addplot[meadowcolor] table [x=t, y=meadows, col sep=comma] {images/example/proba.csv};
	\addplot[wbarleycolor, thick] table [x=t, y=winter barley, col sep=comma] {images/example/proba.csv};
	\addplot[corncolor] table [x=t, y=corn, col sep=comma] {images/example/proba.csv};
	\addplot[wheatcolor,] table [x=t, y=winter wheat, col sep=comma] {images/example/proba.csv};
	\addplot[sbarleycolor] table [x=t, y=summer barley, col sep=comma] {images/example/proba.csv};
	\addplot[clovercolor] table [x=t, y=clover, col sep=comma] {images/example/proba.csv};
	\addplot[triticalecolor] table [x=t, y=winter triticale, col sep=comma] {images/example/proba.csv};
	
	\draw[fill=white, draw=none, opacity=.5] (axis cs:\tstopv,0) rectangle (axis cs:1,1);
	
	\draw (axis cs:\tstopv,0) -- (axis cs:\tstopv,2.2);
	
	\node[annot](rand) at (axis cs:.1,.8) {first hints};
	\draw[point] (rand) -- (axis cs:.2,.15);
	
	\node[annot](wrong) at (axis cs:.17,1.1) {initially wrong predictions};
	\draw[point] (wrong) -- (axis cs:.25,.7);
	\draw[point] (wrong) -- (axis cs:.4,1);
	
	\node[annot, align=center](corrwait) at (axis cs:.4,1.3) {waiting for more data};
	\draw[point] (corrwait) -- (axis cs:.5,.8);
	
	\node[annot](corr) at (axis cs:.8,1.3) {seen enough data};
	\draw[point] (corr) -- (axis cs:.63,1);
	
	\legend{meadows,\textbf{winter barley},corn,winter wheat,summer barley,clover,winter triticale}


	\end{groupplot}
	
	\node[font=\sffamily\tiny, anchor=west] at (-.4,3.3) {All 13 spectral bands of the Sentinel 2 satellites grouped by spectral signature};
	\node[font=\sffamily\tiny, anchor=west] at (0,-3.1) {predicted crop type classes (correct class in \textbf{bold})};
	
	\end{tikzpicture}
		\caption{
			An example of the predictions of the early classification mechanism $\yhat, \deltat = f(\xuptot)$.
			A binary stopping decision $\tstop$ is sampled from the evaluated stopping probability $\deltat$. 
			The upper illustration shows the \emph{Sentinel 2} time series $\V{x}$ based on which the class probabilities $\yhat$ and the stopping time $\tstop$ are produced.}
		\label{fig:introexample}
	\end{figure}
	
	The identification of crop types form space-born imagery forms an important component of agricultural monitoring at a global scale.
	Information obtained from satellite data is often the only source of information to assess the state national agriculture.
	This is particularly important in developing countries where agricultural production is rarely monitored or controlled by central authorities.
	Crucial information on the effect of droughts and shortages are typically available after the end of the vegetative periods which may be too late for effective countermeasures.
	Obtaining knowledge about the expected crop type as early as possible with space-born crop monitoring can play a significant role in providing food security, preventing famine, and determining policies for sustainable agriculture.
	Monitoring of vegetation has been an increasing focus of Earth observation programs. 
	NASA's Landsat or ESA's Sentinel satellites observe the surface of the Earth at weekly intervals at spatial resolutions of 10-60m.
	This allows for regular observations of comparatively small objects, such as field parcels. 
	While enormous quantities of data are collected, methods to extract information, \eg \citet{torbick2018fusion,chen2018mapping, russwurm2018multi}, are often applied at later times when the vegetative period has ended to ensure a certain accuracy of the prediction.
	With the proposed early classification mechanism, we enable a generic multi-temporal classification model to choose the time of classification $\tstop$ sampled from a probability $\deltat$. 
	Our results show empirically that the classification stopped when sufficient information has been observed to justify a confident classification, as shown in an example in \cref{fig:introexample}.
	
	\section{Related Work}
	
	Multi-temporal approaches commonly rely on the classification of the entire time series either using features extracted from expert knowledge~\citep{bailly2016classification} or data-driven end-to-end learning~\citep{sharma2018land,interdonato2019duplo,russwurm2018multi,pelletier2019}.
	Especially for seasonally variable classes, such as vegetation monitoring or crop type mapping, region-specific expert knowledge is used to restrain the period of the time series to the period where the observed classes are deemed observable, \ie the growing season~\citep{torbick2018fusion,chen2018mapping}. 
	%
	Early classification is the task of predicting the class of an incoming time series $\M{X} = (\V{x}_0, \V{x}_1, \dots , \V{x}_T)$ of observations $\V{x}_t$ as early as possible.
	Common approaches formalize a mixed cost function that could take into account both earliness and accuracy, and optimizing this cost function or a surrogate of it~\cite{dachraoui2015early,tavenard2016cost,mori2017early}.
	Recently, an end-to-end fine-tuning framework has been proposed~\citep{russwurm2019end} to jointly learn 
	\begin{enumerate*}
		\item an embedding for partial time series,
		\item a classifier and
		\item a stopping rule, where both the classifier and the stopping rule are computed from the embedding.
	\end{enumerate*}
	We will use this formulation in this work for the use-case of agricultural monitoring with a novel loss function that allows jointly optimizing parameters for classification model and stopping decision in one training phase.
	
	\section{Methodology}
	\label{sec:method}
	
	Our model estimates classification scores $\yhat_t$ and a probability of stopping $\deltat$ at each time $t$. 
	It is composed of a feature extraction network whose output is $\V{h}_t = \ffeat(\xuptot)$ (\cref{sec:classmodel}) and two light-weight mappings for classification scores $\yhat_t  = \fclass(\V{h}_t)$ and stopping probability $\deltat = \fdelta(\V{h}_t)$ (\cref{sec:mechanism}).
	We denote vectors and matrices by bold face and differentiate them by case.
	A dataset $\mathcal{D}=\{(\M{X}^i,\V{y}^i)\}_{i=1}^N$
	with single example $\M{X}^i \in \R^{\{T\times D\}}$ is composed of $T$ $D$-dimensional observations where the subscript is used to indicate time $t$. 
	Temporal sequences up to time $t$ is denoted by an arrow, \eg $\xuptot$, while prediction scores are represented as vector $\yhat \in \R^{M}$ of $M$ classes with the prediction score, \ie the probability output by the model.
	The probability of the correct class is denoted as $\ycorrect$.

	\subsection{Early Classification Mechanism}
	\label{sec:mechanism}
	
	First, we describe the temporal classification model that evaluates a classification score $\yhat_t = \fclass(\ffeat(\xuptot))$ at each time $t$.
	This hidden feature vector $\V{h}_t = \ffeat(\xuptot)$ is typically obtained as a latent representation after a series of cascaded layers in deep classification models.
	Such models produce classification scores by applying a non-linear mapping $\yhat_t = \fclass(\V{h}_t) = \text{softmax}\left(\thetaclass\V{h}_t + \biasclass\right)$.
	Since this hidden state contains information of the predicted class it encodes an internal representation of the classification confidence.
	So, it can be potentially used as an input for the decision of stopping the classification if enough data has been observed.
	A loss function $\mathcal{L}_\text{c} (\xuptot, \V{y})$ evaluates the quality of the prediction compared to ground truth labels $\V{y}$.
	Gradients of this loss are then back-propagated to adjust the classification model parameters $\thetaclass \in \R^{H \times M}, \biasclass \in \R^M$.
	The number of hidden states $H$ is a hyperparameter of the model while $M$ denotes the number of classes.
	
	The early classification mechanism augments this classification model by an additional non-linear mapping $\deltat = \fdelta(\V{h}_t) = \sigma(\thetadelta\V{h}_t + \biasdelta)$ with sigmoidal activation function $\sigma(\cdot)$ that produces a one-dimensional probability of stopping $\deltat$.
	The parameterization $\thetadelta \in \R^{H \times 1}, \biasdelta$ of this mapping, however, can not be adjusted from gradients of the classification loss $\mathcal{L}_c$ directly.
	Hence, a new loss function $\mathcal{L}_t(\xuptot, \V{y})$ is required to adjust both $\thetadelta,\biasdelta$ and $\thetaclass,\biasclass$.
	
	This earliness-aware loss function is then weighted by the probability of stopping 
	\begin{equation}
	P(t;\deltauptot) = \deltat \cdot \prod_{\tau=0}^{t-1} 1 - p_\tau \, .
	\label{eq:pts}
	\end{equation}
	at corresponding time $t$.
	To ensure that this probability will sum to one, $\delta_{t=T}$ is set to 1 regardless of the last hidden state $\V{h}_T$.
	This mechanism utilizes information about the full sequence at training time to restrain $P(t;\deltauptot)$ and calculate the training loss
	\begin{equation}
	\mathcal{L}(\V{x}, \V{y}) = \sum_{t=0}^T P(t;\deltauptot) \mathcal{L}_t(\xuptot, \V{y}) \, .
	\label{eq:ptloss}
	\end{equation}
	
	At inference time when $P(t;\deltauptot)$ cannot be normalized to a probability, $\deltat$ is used to sample a stopping time $\tstop$.
	The prediction $\V{y}_{\tstop}$ at this time is used to derive a classification label for the time series and to assess the accuracy.
	Note that we initialize the weights so that $\deltat \approx 0$ to favor late observations at the beginning of the training period.
	
	\subsection{Earliness Reward Loss Function}
	\label{sec:earlynessreward}
	
	
	Inspired by existing works \cite{mori2017early, tavenard2016cost, russwurm2019end}, we employ a loss 
	\begin{equation}
	\mathcal{L}_t(\xuptot, \V{y} ; \alpha) = \alpha \mathcal{L}_c (\xuptot, \V{y}) - (1 - \alpha)\mathcal{R}_e(t, \ycorrect_t)
	\label{eq:earlyrewardloss}
	\end{equation} 
	$\alpha$-weights a classification loss $\mathcal{L}_c (\xuptot, y)$  with a \emph{earliness reward} term
	\begin{equation}
	\mathcal{R}_e(t, \ycorrect_t) = \ycorrect_t \left(1 - \frac{t}{T}\right) \,.
	\label{eq:earlyrewardterm}
	\end{equation}
	This term linearly scales the potential reward for earlier classifications $\left(1 - \frac{t}{T}\right)$.
	We remind that $\yhat_t$ is the prediction of the network at time $t$ and that $\ycorrect_t$ is the probability of the annotated class.
	This rewards the classifier only for an early classification if the correct class has been predicted at a reasonable accuracy.
	By combining \cref{eq:ptloss,eq:earlyrewardloss,eq:earlyrewardterm} we see that three terms influence the earliness reward.
	These are the probability $P(t;\deltauptot)$, the evaluated classification score of the correct class $\ycorrect_t$ and a linear scaling term $\left(1 - \frac{t}{T}\right)$ that rewards earlier classifications.
	
	\subsection{The Early Classification LSTM Network}
	\label{sec:classmodel}
	
	\begin{figure}
		\centering\tikzstyle{dummy} = [inner sep=0]
\tikzstyle{flow} = [thin, -{Stealth[scale=.5]}]
\tikzstyle{endflow} = [flow, shorten >= 0, shorten <= 0]
\tikzstyle{operator} = [inner sep=0, font=\scriptsize]
\tikzstyle{conn} = [-stealth, shorten >= .2em, inner sep=0]
\tikzstyle{conntime} = [conn, tumgray]
\tikzstyle{lstmcell} = [inner sep=0, fill=tumbluelight!20, rounded corners=1em]

\newcommand{\act}{
	\begin{tikzpicture}[scale=.3]
		\node[circle, draw]{
			\begin{tikzpicture}
			\draw (0,0) to[out=0, in=180] (1,1);
			\end{tikzpicture}
		}
	\end{tikzpicture}	
}

\newcommand{\lstm}{
	\begin{tikzpicture}[inner sep=0, xscale=.5, yscale=.5]
	\coordinate (-input) at (0,1); 
	\coordinate (-output) at (0,-2.75); 
	\node[inner sep=0](fgate) at (-1.5,0){\fcn};
	\node[inner sep=0](igate) at (-.5,0){\fcn};
	\node[inner sep=0](jgate) at (.5,0){\fcn};
	\node[operator](jmult) at (0,-1.25) {$ \odot $};
	\draw[endflow] (jgate) -- (jmult);
	\draw[endflow] (igate) -- (jmult);
	\node[inner sep=0](ogate) at (1.5,0){\fcn};

	\draw[endflow] (-input) to[out=270,in=90] (ogate.north);
	\draw[endflow] (-input) to[out=270,in=90] (jgate.north);
	\draw[endflow] (-input) to[out=270,in=90] (igate.north); 
	\draw[endflow] (-input) to[out=270,in=90] (fgate.north);

	\node[operator](fmult) at (-1,-1.25) {$ \odot $};
	\draw[endflow] (fgate) -- (fmult); 
	\node[operator](cadd) at (0,-1.75) {$\oplus$};
	\draw[endflow] (jmult) -- (cadd); 
	\draw[endflow] (fmult) -- (cadd);		
	
	\node[operator](outtanh) at (1,-1.25) {$\odot$};
	\draw[endflow] (cadd) -- (outtanh);
	\draw[endflow] (ogate) -- (outtanh);
	
	\node[font=\tiny](c) at (-1,-2){$\V{c}_t$};
	\draw[endflow] (c) -- (fmult);
	\draw[endflow, tumgray] (cadd) -- (c);
	
	\draw (outtanh) to[in=90, out=270] (-output);
	\end{tikzpicture}
}

\newcommand{\legend}{
	\begin{tikzpicture}[yscale=.8, font=\scriptsize]
		\node[label distance=-1em, label={above:fully connected}] at (0,0){$\fcn = \sigma\left(\M{\theta}\V{x}+\V{b}\right)$};
		\node[label={above:hidden state}] at (0,1){$\hidden{6}: \V{a} \in \mathbb{R}^h$};
		\node[label={above:observed state}] at (0,2){$\drawvector{6}: \V{a} \in \mathbb{R}^n$};
	\end{tikzpicture}
}

\tikzsetnextfilename{lstmmodel}
\begin{tikzpicture}[node distance=1em and 3em, font=\sffamily]
\node[fill=tumgraylight!20, rounded corners, inner sep=1pt](legend) at (3,-2.5){\legend};

\node[label={left:$\V{x}_t$}](x0){\drawvector{13}};
\node[norm, below= .7em of x0](nx0){\scriptsize LayerNorm};
\node[lstmcell, below=of nx0](l0){\lstm};
\node[norm, below=of l0](ny0){\scriptsize LayerNorm};
\node[below=.7em of ny0, label=left:$\V{h}_t$](h){\hidden{16}};

\node[left=2em of l0](ll0){$L \times$};

\node[below left= 2em and 1.3em of h, label={left:$\delta_t$}](d){\drawvector{1}};
\node[below right= 2em and 1em of h, label={right:$\yhat_t$}](y){\drawvector{6}};
\draw (h) -- node[fill=white, inner sep=2pt, label={right:}]{\fcn} (y);
\draw (h) -- node[fill=white, inner sep=2pt, label={left:}]{\fcn} (d);

\draw[conn] (x0) -- (nx0);
\draw[conn] (nx0) -- node[fill=white]{\drawvector{13}} (l0);
\draw[conn] (l0) -- node[fill=white]{\hidden{16}}(ny0);
\draw[conn] (ny0) -- (h);

\draw[conntime] (l0)++(-4em,0) -- node[midway, above, font=\tiny]{$\V{h}_{t-1}$} (l0);

\draw[conntime] (l0) -- ++(4em,0);

%
%
%
%
%
%

\end{tikzpicture}
		\caption{Schematic illustration of the multi-layer long short-term memory classification model equaipped with tho linear mappings for the class probabilities $\yhat$ and the stopping probability $\deltat$.}
		\label{fig:network}
	\end{figure}
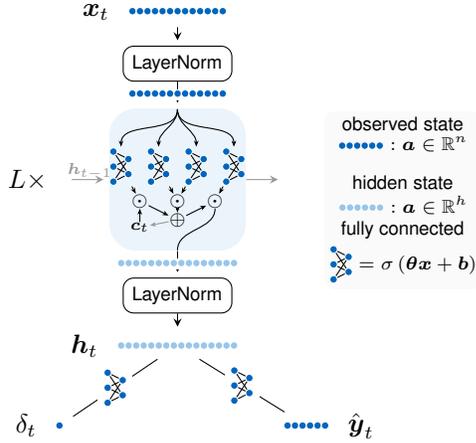
	
	To test this early classification mechanism on a real-world remote sensing vegetation dataset, we implemented a light-weight multi-layer \emph{long short-term memory (LSTM)} \citep{hochreiter1997long} recurrent network, as this architecture has shown good results in similar use-cases in recent studies \cite{russwurm2018multi, interdonato2019duplo, sharma2018land}. 
	\Cref{fig:network} shows a detailed representation of the employed gated recurrent neural network architecture with the previously described early classification mechanism.
	The input vector $\V{x}_t \in \R^d$ at time $t$ is first $z$-normalized by mean and variance parameters using \emph{layer normalization} \citep{ba2016layer}, which are computed from the training set.
	The normalized representation is then provided to $L$ cascaded LSTM layers $\V{h}_t^{l} = \text{lstm}^l(\V{h}_t^{l-1}, \V{h}_{t-1}^{l})$.
	We denote the $l$-th layer in the network as a raised index.
	Each of these gated recurrent layers evaluates a feature representation $\V{h}_t^{l}$ based on two hidden representations:
	one at the current time of the previous layer $\V{h}_t^{l-1}$ and one of the same layer at the previous time step $\V{h}_{t-1}^{l}$.
	Long short-term memory cells additionally utilize an internal cell state $\V{c}_t^{l}$ to store long-term information, as indicated by the circular connection in the illustration.
	The last layer-normalized hidden representation $\V{h}_t$ then computes class prediction scores $\hat{\V{y}}$ for each class and the stopping probability $\deltat$, as described in \cref{sec:mechanism}.
	
	\section{Results and Experiments}
	
    To test the early classification mechanism, we obtained the label and geometry information of 40k field parcels from local authorities of the 2018 season in central Europe.
	Over this region we acquired \emph{all} available \emph{Sentinel 2} images from \emph{Google Earth Engine} \citep{gorelick2017google}.
	All $D=13$ spectral bands located within one field parcel were mean-aggregated to a feature vector $\V{x}_t$, as shown in \cref{fig:introexample}.
	From the available satellite observations, we randomly choose $T=70$ times to introduce a variance to the data samples and prevent overfitting.
	Geographic parcels are known to be spatially autocorrelated due to their proximity to each other \cite{tobler1970computer}.
	Hence, we avoid a field wise random partitioning to training, validation and evaluation partitions.
	Instead, we group the area of interest first in spatially separate regions and then assign all fields within one region to the respective partition, as suggested by \citet{jean2018tile2vec, russwurm2018multi}.
	
	We determine a set of optimal hyperparameters for the classification model by training  without earliness mechanism on standard cross-entropy loss $\mathcal{L}(\V{x},\V{y}) = \frac{1}{T}\sum_{t=0}^T -\log(\ycorrect)$ with equal weight over all observations.
	For the weight updates, we used the \emph{Adam optimizer} \citep{kingma2014adam} and dropped-out connections between recurrent layers at 50\% probability and a batch size of 1024.
	We searched over the number of recurrent layers $L \in \{1,2,3\}$, the size of the hidden dimension $H \in \{64, 128, 256, 512\}$, and the learning rate $\nu \in \{0.1, 0.01, 0.001\}$ for 30 epochs.
	The parameters $L=4, H=64, \nu=0.01$ that achieved best {accuracy} on the {validation} were used throughout this work.
	When training the model from scratch, we favor initially late classifications with $\deltat \approx 0$.
	To achieve this, we initialize the bias term $\deltat$ with a negative non-zero mean normal distribution $\biasdelta \sim \mathcal{N} (\mu=-0.2,\sigma=0.1)$.
	The training of one model took approximately ten minutes on a \textsc{\small Nvidia GeForce GTX 1060}.
		
	\subsection{Qualitative Example}
	
	We illustrated the early classification results qualitatively in \cref{fig:introexample} on a classification example using raw Sentinel 2 top-of-atmosphere reflectance data gathered over the year 2018.
	In this multi-temporal classification task, we estimate classification scores $\yhat_t$ for each evaluated class based on observed data $\xuptot$ at each time $t$. 
	In the example, we can see that the model initially predicts randomly from the January to March period.
	Then some classification-relevant features are observed, which initially mislead the classifier to false predictions in the April period.
	Finally, the correct class is assigned the highest prediction score from May onwards.
	Early classification augments this classification scheme by estimating a stopping probability $\deltat$ out of which a stopping time $\tstop$ is sampled.
	This stopping decision serves as an indication if a sufficient amount of data has been observed and no further data is required to allow for a confident classification.
	
	\subsection{Quantitative Accuracy vs Earliness Analysis}
	
	\begin{table}
		
		\scriptsize
		\hspace{0em}\begin{tabular}{lcccccc}
			\toprule\small
			\textbf{$\alpha$} & accuracy & $\meantstop$  & precision & recall & $f_1$ & $\kappa$ \\
			\cmidrule(lr){0-0}\cmidrule(lr){1-1}\cmidrule(lr){2-2}\cmidrule(lr){3-3}\cmidrule(lr){4-4}\cmidrule(lr){5-5}\cmidrule(lr){6-6}\cmidrule(lr){7-7}
			.0 & .25 $\pm$ .22 & .10 $\pm$ .17 & .19 $\pm$ .20 & .25 $\pm$ .17 & .16 $\pm$ .20 & .12 $\pm$ .19 \\
			.2 & .81 $\pm$ .03 & .40 $\pm$ .02 & .70 $\pm$ .01 & .74 $\pm$ .01 & .71 $\pm$ .01 & .71 $\pm$ .04 \\
			.4 & .80 $\pm$ .09 & .47 $\pm$ .03 & .71 $\pm$ .02 & .74 $\pm$ .01 & .71 $\pm$ .02 & .71 $\pm$ .10 \\
			.6 & .85 $\pm$ .02 & .88 $\pm$ .07 & .73 $\pm$ .04 & .74 $\pm$ .03 & .73 $\pm$ .03 & .77 $\pm$ .03 \\
			.8 & .84 $\pm$ .01 & .93 $\pm$ .05 & .72 $\pm$ .02 & .75 $\pm$ .01 & .73 $\pm$ .02 & .76 $\pm$ .02 \\
			1.0 & .83 $\pm$ .03 & 1.00 $\pm$ .00 & .72 $\pm$ .03 & .75 $\pm$ .01 & .72 $\pm$ .03 & .75 $\pm$ .04 \\
			\bottomrule
		\end{tabular}
		\caption{Varying the weighting factor $\alpha$ for \emph{early reward} loss formulation (\cref{sec:earlynessreward}).}
		\label{tab:alpha}
		
	\end{table}
	
	In \cref{tab:alpha}, we quantitatively evaluated the effect of the trade-off parameter $\alpha$ on the classification performance on a series of accuracy metrics.
	We trained models with different weight initializations and batching sequences three times to assess the stability of the training process under a range of trade-off parameters $\alpha \in \{0.2,0.4,0.6,0.8\}$.
	For each $\alpha$ we report a series of accuracy metrics and the averaged time of classification $\meantstop = \frac{1}{N} \sum_{i=0}^{N} \tstop$ normalized as fraction of the total sequence length.
	The \emph{overall accuracy} describes the parcel-wise ratio of correctly positive and correctly negative examples in relation to the total number of samples.
	Precision, recall and their harmonic mean $f1$-score are calculated on a class-wise basis which is less sensitive to the frequency of samples per classes compared to {overall accuracy} or the mean time of classification $\meantstop$.
	We also report the kappa metric $\kappa$~\citep{fleiss1969large}, as it is popular in the remote sensing community and similarly accounts for imbalanced class distributions by normalizing the score with a probability of random prediction.
	The early reward loss function was in terms of achieved accuracy resilient to the choice of $\alpha$.
	This becomes apparent as the classification accuracies remained stable throughout the evaluated ranges of $\alpha$.
	
	\subsection{Class-wise Analysis of Stopping Times}
	\label{sec:classes}
	
	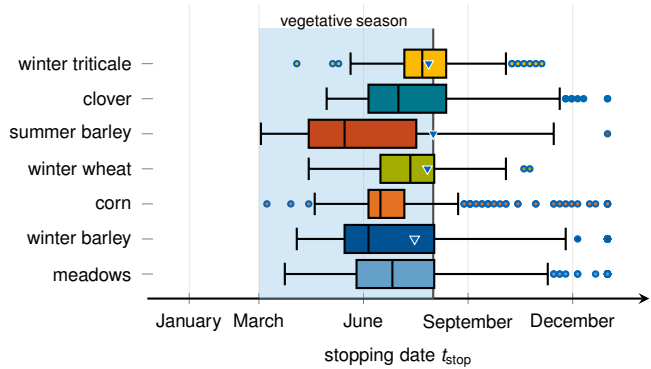
\begin{figure}
		\tikzsetnextfilename{classboxplots}
\begin{tikzpicture}

\tikzstyle{boxstyle}=[
	mark options={
	draw=tumblue,
	scale=0.5},
	mark=*,
	solid,
	draw=black]

\begin{axis}[
ytick={1,2,3,4,5,6,7},
yticklabels={
	meadows,
	winter barley,
	corn,
	winter wheat,
	summer barley,
	clover,
	winter triticale},
xmajorgrids,
height=5.5cm,
xmin = 0,
xmax = 1,
ymin = 1,
ymax = 8,
width=\linewidth,
y axis line style={draw=none},
xtick={0,0.1666666667,0.4166666667,0.6666666667,0.9166666667},
xticklabels={January,March,June,September,December},
xlabel={stopping date $\tstop$}
]

\draw [fill=tumbluelight, opacity=.4, draw=none] (axis cs:0.1666666667,0) rectangle (axis cs:0.5833333333,8);
\draw[draw=tumgraydark] (axis cs:0.5833333333,0) -- (axis cs:0.5833333333,8);

\node[font=\tiny\sffamily] at (axis cs:0.37,8.2){vegetative season};


\addplot+[boxplot, fill=meadowcolor, boxstyle] table[x = meadows, col sep=comma]{images/logs/data/early_reward_p2/classes/meadows.csv};
%

\addplot+[boxplot, fill=wbarleycolor, boxstyle] table[x=winter barley, col sep=comma]{images/logs/data/early_reward_p2/classes/winter_barley.csv};

%

\addplot+[boxplot, fill=corncolor, boxstyle] table[x =corn, col sep=comma]{images/logs/data/early_reward_p2/classes/corn.csv};
%

\addplot+[boxplot, fill=wheatcolor, boxstyle] table[x =winter wheat, col sep=comma]{images/logs/data/early_reward_p2/classes/winter_wheat.csv};

%

\addplot+[boxplot, fill=sbarleycolor, boxstyle] table[x =summer barley, col sep=comma]{images/logs/data/early_reward_p2/classes/summer_barley.csv};


\addplot+[boxplot, fill=clovercolor, boxstyle] table[x=clover, col sep=comma]{images/logs/data/early_reward_p2/classes/clover.csv};


\addplot+[boxplot, fill=triticalecolor, boxstyle] table[x=winter triticale, col sep=comma]{images/logs/data/early_reward_p2/classes/winter_triticale.csv};

\def\triticale{0.5722222222}
\def\sbarley{0.5833333333}
\def\wbarley{0.5388888889}
\def\wheat{0.5694444444}

\node[druschdatum] at (axis cs:\triticale,7){};

\node[druschdatum] at (axis cs:\sbarley,5){};

\node[druschdatum] at (axis cs:\wheat,4){};

\node[druschdatum] at (axis cs:\wbarley,2){};


\end{axis}
\end{tikzpicture} 
		\caption{
			Plot of distributions of the stopping times grouped by individual classes. The stopping times for each class follows a different distribution. 
			The prediction for some classes, \eg \emph{barley}, are on median earlier than the others, \eg \emph{corn} or \emph{wheat}, that are stopped at a later.
			The evaluated stopping times are correlated with actual phenological information, such as the vegetative season (shaded in blue) or the date of harvest (\druschdatum) provided by local authorities.
		}
		\label{fig:boxplots}
	\end{figure}
	
	Next, we examine the stopping times per class and evaluate if these are related to phenologically characteristic events of the crop classes.
	We show the distribution of stopping times of all field parcels per individual class as boxplots in \cref{fig:boxplots}. 
	The distributions over stopping times vary for each of the respective classes. 
	For instance, the quartiles of the \emph{corn} distribution is narrow which means that 50\% of the observed parcels are identified within one to two weeks, while crop types, such as \emph{barley} or \emph{meadows} are spread over one month.
	One can observe that the stopping times generally correlate with the vegetative period (shaded in blue) and the harvest times (marked by triangle \druschdatum).
	We obtained these dates from local authorities to be able to assess the performance of the early classification mechanism objectively.
	Note, however, that no region-specific expert knowledge is required for evaluating the stopping probabilities. Hence, the shown correlation originates solely from the observed satellite data which can be obtained globally. 
	Overall, we can observe that our early classification model decided to stop the classification before the end of the vegetative period for a majority of field parcels.
	These results demonstrate the potential of early classification for agricultural monitoring where a confident classification decision can be determined independently from local expert knowledge and often before the end of the vegetative season.

	\section{Conclusion}

    In this work, we introduced a novel end-to-end trainable early classification mechanism for satellite-based agricultural monitoring.
	It can augment any multi-temporal classification models that evaluate a hidden representation at each time.
	Further, we analyzed the stopping times for the individual vegetation classes.
	Here, we saw that the parametrization of the stopping mechanism for each class was learned at different times and that the final times of stopping followed distinct distributions from crop type to crop type.
	We compared these evaluated stopping times with phenological and ontological information from local authorities and saw that many field parcels could be classified before the end of the vegetative period which is the usual end time of classification for regular crop type mapping approaches.
	
	{\small
		\bibliography{bib/references}
		\bibliographystyle{icml2019}
	}

\end{document}